\documentclass[11pt]{article}

\PassOptionsToPackage{table}{xcolor}
\usepackage[final]{acl}

\usepackage{times}
\usepackage{latexsym}

\usepackage[T1]{fontenc}

\usepackage[utf8]{inputenc}

\usepackage{microtype}

\usepackage{inconsolata}

\usepackage{graphicx}

\usepackage{booktabs}
\usepackage{tabularx}
\usepackage{makecell}

\usepackage{times}
\usepackage{latexsym}
\usepackage{caption}
\usepackage{subcaption}
\usepackage{amsmath}
\usepackage{amssymb}
\usepackage{amsfonts}
\usepackage{subfloat}
\usepackage{xcolor}
\usepackage{siunitx}
\usepackage{cleveref}

\usepackage{tcolorbox}
\tcbuselibrary{breakable}

\definecolor{color1}{HTML}{7676f9}
\definecolor{color2}{HTML}{c9c9f4}
\definecolor{color3}{HTML}{e6e6f3}
\definecolor{color4}{HTML}{e6e6f3}
\definecolor{color5}{HTML}{e8e8f3}
\definecolor{color6}{HTML}{e8e8f3}
\definecolor{color7}{HTML}{e9e9f2}
\definecolor{color8}{HTML}{efeff2}
\definecolor{color9}{HTML}{f1f1f2}
\definecolor{color10}{HTML}{f2f1f1}

\usepackage{enumitem}

\usepackage{placeins}

\usepackage{url}

\newcommand{\camerareadytext}[1]{\xspace}

\newcommand{\sref}[1]{\S\ref{#1}}
\newcommand{\fref}[1]{Figure~\ref{#1}}
\newcommand{\tref}[1]{Table~\ref{#1}}

\title{What Makes a Good Response? An Empirical Analysis of Quality in Qualitative Interviews}

\author{Jonathan Ivey \\
  Johns Hopkins University  \\
    \texttt{jivey6@jhu.edu} \\\And
  Anjalie Field \\
  Johns Hopkins University \\
  \texttt{anjalief@jhu.edu} \\\And
  Ziang Xiao \\
  Johns Hopkins University \\
  \texttt{ziang.xiao@jhu.edu} \\}

\usepackage{ragged2e}
\begin{document}
\maketitle
\begin{abstract}
Qualitative interviews provide essential insights into human experiences when they elicit high-quality responses. While qualitative and NLP researchers have proposed various measures of interview quality, these measures lack validation that high-scoring responses actually contribute to the study's goals. In this work, we identify, implement, and evaluate 10 proposed measures of interview response quality to determine which are actually predictive of a response's contribution to the study findings. To conduct our analysis, we introduce the Qualitative Interview Corpus, a newly constructed dataset of 343 interview transcripts with 16,940 participant responses from 14 real research projects. We find that direct relevance to a key research question is the strongest predictor of response quality. We additionally find that two measures commonly used to evaluate NLP interview systems, clarity and surprisal-based informativeness, are not predictive of response quality. Our work provides analytic insights and grounded, scalable metrics to inform the design of qualitative studies and the evaluation of automated interview systems.\footnote{\RaggedRight Dataset: \url{https://doi.org/10.5064/F6JWVCH6} \\ Code: \url{https://github.com/jonathanivey/interview-quality}}

\end{abstract}

\section{Introduction}

Qualitative interviews are a primary method for surfacing insights into experiences, motivations, and behaviors that quantitative methods cannot capture. However, the value of the insights an interview produces depends directly on the quality of the responses it elicits, and our understanding of what makes a response high-quality rests almost entirely on theoretical intuition.
Qualitative researchers have proposed characteristics of high-quality interview responses, such as spontaneity and relevance \citep{kvale_interviews_2009, charmaz_constructing_2014, patton_qualitative_2015, small_qualitative_2022}, but these frameworks disagree substantially on which characteristics matter, and none offer empirical evidence that responses with these characteristics actually contribute to a study's findings.
Such evidence is necessary for determining which measures should guide interview practices.

Recent interest in AI has accelerated the need to understand and quantify interview data quality. NLP systems are increasingly being used to conduct or assist human interviews. For example, Anthropic recently deployed a system to autonomously collect qualitative responses to investigate how professionals use AI \citep{handa_introducing_2025}. Other applications include academic research \citep{liu_envisioning_2025}, market research \citep{anugraha_sparkme_2026}, preference elicitation \citep{choudhury_bed-llm_2025}, and gathering public feedback \citep{jiang_communitybots_2023}.
Current interview systems commonly use proxy criteria for judging elicited response quality like specificity, clarity, and relevance \citep{xiao_if_2020, xiao_tell_2020, jiang_communitybots_2023, hu_designing_2024, jacobsen_chatbots_2025}, but these measures similarly lack validation that high-scoring responses contribute to study findings. Without validated evaluation metrics, building and evaluating AI systems for qualitative research remains untenable.

In this work, we investigate characteristics of interview response quality through the identification and implementation of proposed quality metrics and empirical analysis of a new dataset. First, we identify 10 measures of interview response quality through a review of qualitative literature and research studies on NLP interview systems. We empirically assess these 10 measures over a newly constructed dataset of 343 transcripts from 14 real qualitative research studies. Our analysis of 16,940 participant responses reveals which measures are actually predictive of a response's contribution to the study findings, our criterion for overall quality. 

From our analysis, we find that the measure most predictive of response quality is relevance to a key research question of the study. We also find that responses containing the kind of insights unique to qualitative studies are more likely to be high quality, for example, responses that explain why a belief or experience matters personally to the participant. Finally, we find that two measures commonly used to evaluate interviewer systems, clarity and surprisal-based informativeness, are not significantly predictive of response quality. As the end goal of quality measures is to inform interview strategies, we further use our measures to conduct a case study of how time and interview techniques affect response quality.

Our contributions in this work include (1) the Qualitative Interview Corpus, a newly constructed dataset of 343 transcripts from 14 qualitative research projects that enables empirical analysis of qualitative interviews, (2) the creation and validation of automated measures of qualitative interview characteristics, (3) empirical analysis of which characteristics of participant responses are predictive of overall response quality, and (4) an example use case of how these measures can inform interview strategies.
Our work offers the first empirical analysis of response characteristics in qualitative interviews, offering grounded metrics that can inform both the design of qualitative studies and the evaluation of NLP interview systems.

\section{Methods}

\begin{table*}[t]
\small
\centering
\begin{tabularx}{\textwidth}{@{} l >{\hsize=1.1\hsize}X >{\hsize=.9\hsize}X @{} }
\toprule
\textbf{Characteristic} & \textbf{Definition} & \textbf{Source} \\ \midrule
\makecell[tl]{Specificity \\ (Palpability)} & The extent to which a response provides detailed examples rather than abstract generalizations. & \citet{kvale_interviews_2009, charmaz_constructing_2014, xiao_tell_2020, small_qualitative_2022} \\ \midrule
Clarity & How clear and understandable a response is. & \citet{kvale_interviews_2009, xiao_tell_2020} \\ \midrule
Immediate Relevance & How relevant the response is to the specific question asked by the interviewer. & \citet{patton_qualitative_2015, xiao_tell_2020} \\ \midrule
\makecell[tl]{Research Question \\ Relevance} & How relevant the response is to the overall research question. & \citet{kvale_interviews_2009, charmaz_constructing_2014, patton_qualitative_2015} \\ \midrule
Spontaneity & The extent to which the response provides information beyond what is provided in the question. & \citet{kvale_interviews_2009} \\ \midrule
Self-reportedness & How understandable a response is if taken out of context. & \citet{kvale_interviews_2009} \\ \midrule
\makecell[tl]{Attributed Meaning \\ (Cognitive Empathy)} & The extent to which a response demonstrates the personal significance of a belief or experience to the participant. & \citet{small_qualitative_2022, charmaz_constructing_2014} \\ \midrule
Average Surprisal & The average word-level surprisal of the response. & \citet{xiao_tell_2020} \\ \midrule
\makecell[tl]{Response Length \\ Ratio}& Ratio of the length of the participant response to the length of the interviewer question & \citet{kvale_interviews_2009} \\ \midrule
Response Length & Length of the response. & \citet{xiao_tell_2020} \\ \bottomrule
\end{tabularx}
\caption{We identify 10 key characteristics of interview responses from qualitative literature and NLP interview systems, shown here with their definitions and where they were proposed.}
\label{tab:characteristics}
\end{table*}

To investigate which response characteristics are indicative of their contribution to the study findings, we identify 10 proposed quality measures from qualitative research literature and evaluations of interview systems. We then design a quality criterion based on the extent to which a response contributes to the study findings. Finally, we create an automated measure of these response characteristics and our quality criterion to enable large-scale empirical analysis.

\subsection{Proposed Characteristics of High-Quality Responses}
\label{sec:proposed-characteristics}
We review qualitative literature and evaluations of NLP interview systems to identify the characteristics of participant responses that are commonly used as quality metrics.

In qualitative literature, \citet{kvale_interviews_2009} propose the most robust set of measures, including richness, relevance to the research question, spontaneity, self-reportedness, and the ratio of the length of the participant utterance to the length of the interviewer utterance. \citet{patton_qualitative_2015} further identifies relevance to the research question and relevance to the exact question asked by the interviewer as key aspects of interview quality. \citet{charmaz_constructing_2014} indicates that quality data will be ``rich, substantial, and relevant.'' \citet{small_qualitative_2022} propose an alternative view of qualitative research quality based on five key constructs. Two of these constructs, cognitive empathy and palpability, are characteristics of participant responses. Note that we refer to cognitive empathy as "attributed meaning" to better align with its definition and distinguish it from other characteristics.

In NLP interview systems, the most common measures for response quality are based on Gricean maxims \citep{grice_logic_1975}. This approach identifies quality responses as those with specificity, clarity, relevance, and surprisal-based informativeness \citep{xiao_if_2020, xiao_tell_2020, jiang_communitybots_2023, hu_designing_2024, jacobsen_chatbots_2025}. These characteristics are often considered alongside measures of user engagement such as response length. Other work has combined these with the above measures from qualitative literature \citep{cuevas_collecting_2025}.

To ensure that our final set of characteristics is sufficiently distinct, we identify definitions from each of the original sources and merge characteristics with exceptionally similar definitions, such as specificity and palpability. We choose not to include an explicit measure of richness because, based on the existing definitions, we consider richness to be a combination of other characteristics such as specificity, self-reportedness, and attributed meaning. Finally, to reduce multicollinearity between suprisal-based informativeness and response length, we instead use the average word-level surprisal rather than the total word-level surprisal. The final set of characteristics, definitions, and their origins is outlined in Table \ref{tab:characteristics}.

\subsection{Our Criterion for Response Quality}
\label{sec:quality-criterion}
To identify high-quality responses, we develop a criterion based on \textit{the extent to which a response contributes to the results of a study}. Unlike the previously identified characteristics, our criterion is grounded in research outcomes (i.e., the final results) and cannot be measured during the data collection phase. However, it can be used to compare and validate the other characteristics that can be measured from responses alone, as demonstrated in \sref{sec:regression-analysis}. Our criterion uses the following scoring rubric to estimate the likelihood that a response contributed to the goals of the study:
\begin{enumerate}[noitemsep]
    \item The response is unrelated or contradictory to the results section.
    \item The response is tangentially related to the results section with no specific substance.
    \item The response aligns with the results section but is general or vague.
    \item The response provides an example or sentiment matching the results section's conclusions.
    \item The response appears in the results section or is a primary source for it.
\end{enumerate}

\subsection{Automatically Identifying Response Characteristics}
\label{sec:automating-measures}

To conduct our analysis across a large dataset, we implement automated measures for the 10 response characteristics and our quality criterion. Three of the response characteristics can be computed directly: we compute response length and response length ratio based on the number of tokens, and we compute the average word-level surprisal using \citeposs{oh_frequency_2024} implementation based on token counts from the Pile \citep{gao_pile_2020}.

\paragraph{Conceptual Measures}
\label{sec:conceptual-measures}
The remaining 7 characteristics and our quality criterion require conceptual judgments that we obtain using an LLM judge. For the quality criterion, we prompt the model to rate the participant response on a scale from 1 to 5 according to the rubric in \sref{sec:quality-criterion}. For the other measures, we create rubrics from the definitions in the original sources and use them to prompt the model to rate responses on a scale of 1 to 3. 

In addition to the prompt, we provide the models with (1) the current interview excerpt that the model is rating, (2) the interview excerpt immediately preceding the current excerpt to provide conversational context, and (3) 1--2 sentences providing broad context for how the interviews were conducted and the general goals of the project.

For our quality criterion, we additionally provide the model with a segment of the results section of the paper. For each excerpt $e$, let $S$ be the set of all segments in the results section of the corresponding study, and let $q(e, s)$ represent the estimated likelihood that excerpt $e$ contributed to segment $s \in S$. We evaluate our quality criterion across all segments and take the maximum value to determine the final score $QI(e)$:

$$QI(e) = \max_{s \in S} q(e, s)$$

This single score measures the extent to which a participant response contributed to any of the study results. We use the same process for research question relevance. Letting $Q$ be the set of all key research questions and $r(e, q)$ be the estimated relevance of excerpt $e$ to a single question $q \in Q$, the final relevance score $RQ(e)$ is calculated as:

$$RQ(e) = \max_{q \in Q} r(e, q)$$

The full prompts used for our measures are provided in \Cref{sec:model_prompts}.

\paragraph{Human Validation}
\label{sec:human-validation}
To validate whether LLM judges can estimate these conceptual measures, we compare their outputs using GPT-5.4 and Gemini 3.1 Pro to human judgments on 100 interview excerpts from 5 representative projects in our dataset. For each excerpt, we have three different annotators with experience analyzing qualitative interviews rate the 7 conceptual characteristics and quality criterion for the participant response, resulting in 2,400 total annotations. The 100 excerpts were selected from a random sample that was then balanced to have equal distributions of each rating for each characteristic. We provide annotators with the same information as the LLM with only minor formatting changes, like highlighting participant statements, to reduce cognitive load. An example of the annotation setup is provided in \Cref{sec:annotation_setup}.

\section{Dataset}

\begin{table}[t]
\small
\centering
\begin{tabularx}{\columnwidth}{@{} X c @{}}
\toprule
\textbf{Research Project} & \textbf{\# Interviews} \\ \midrule
Mindfulness for Firefighters and EMS Workers \citep{F68TOJJY_2024} & 11 \\ \midrule
Drug Shortage Management \citep{F6AGWUJG_2021} & 16 \\ \midrule
Ghanaian Healthcare Workers During COVID-19 \citep{F6FYZITI_2025} & 20 \\ \midrule
Socializing Policy Feedback \citep{F6HYTYIJ_2025} & 30 \\ \midrule
Perspectives on Political Representation \citep{F6L9HHYL_2025} & 23 \\ \midrule
Nutrition Interventions in Rural Ethiopia \citep{F6MTPVK7_2025} & 21 \\ \midrule
Marine Corps Education Project \citep{F6AHDRFQ_2020} & 32 \\ \midrule
Intergovernmental Coordination Mechanisms \citep{F6QHVGUI_2023} & 43 \\ \midrule
Models of Delivery for Online Spiritual Care \citep{F6R7J9HL_2025} & 21 \\ \midrule
Partnership between Kidney Disease Patients and Caregivers \citep{F6UXQABW_2024}& 25 \\ \midrule
Shared Data for Learning Qualitative Data Analysis \citep{F6XZV8BZ_2025} & 9 \\ \midrule
Advance Care Planning in Hospice Organizations \citep{F6YMWPUX_2021} & 50 \\ \midrule
Food Retail and Service Workers during COVID-19 \citep{F6Z82KER_2024} & 23 \\ \midrule
High-performance School-age Athletes at Australian Schools \citep{F6ZP448B_2017} & 19\\ \bottomrule
\end{tabularx}
\caption{The Qualitative Interview Corpus is built from 14 research projects across a diverse set of domains. This table lists each project and the number of interviews that it contributed to the corpus.}
\label{tab:data-deposits}
\end{table}

\begin{table*}[t]
\small
\centering
\begin{tabular*}{.75\textwidth}{@{\extracolsep{\fill}} l c c c @{} } %
\toprule
\textbf{Conceptual Measure} & \textbf{\makecell{Human \\ Agreement}} & \textbf{\makecell{GPT-Human \\  Agreement}} & \textbf{\makecell{Gemini-Human \\  Agreement}} \\ \midrule
Attributed Meaning & 0.750 & 0.868 & 0.757 \\ \midrule
Spontaneity & 0.716 & 0.797 & 0.679 \\ \midrule
Specificity & 0.732 & 0.789 & 0.807 \\ \midrule
Immediate Relevance & 0.602 & 0.764 & 0.586 \\ \midrule
Response Quality & 0.754 & 0.757 & 0.696 \\ \midrule
\makecell[tl]{Research Question Relevance} & 0.714 & 0.690 & 0.641 \\ \midrule
Self-reportedness & 0.754 & 0.679 & 0.698 \\ \midrule
Clarity & 0.598 & 0.606 & 0.480 \\ \bottomrule
\end{tabular*}
\caption{From 2,400 human judgments of our conceptual measures, we find comparable agreement between human ratings (Human) and between the median human ratings and LLM judge ratings (GPT-Human and Gemini-Human) as measured with Krippendorff's alpha.}
\label{tab:agreement_results}
\end{table*}

To our knowledge, there is no openly available dataset for analyzing qualitative interviews across multiple domains. To enable empirical analysis of qualitative interviews, we introduce The Qualitative Interview Corpus: a dataset of 343 qualitative interviews and their corresponding papers from 14 research projects across a diverse set of domains (\tref{tab:data-deposits}; see \Cref{sec:descriptive-stats} for more details).

\paragraph {Data Curation} We construct the corpus from deposits to The Qualitative Data Repository,\footnote{\url{https://qdr.syr.edu}} an archive for storing and sharing digital data collected through qualitative and multi-method research. We select data deposits from projects that conducted English qualitative interviews, provided anonymized transcripts, have openly accessible data, and have a corresponding research paper with the results of their study. We manually review the data and exclude projects that do not elicit open-ended participant responses (e.g., surveys that were conducted orally and then transcribed). Our final dataset contains 14 research projects with a total of 343 interviews.

\paragraph{Preprocessing} To make the data suitable for computational analysis, we first extract 58,688 utterances from the PDFs of the interview transcripts. We use speaker tags from the transcripts to assign each utterance to either the participant (31,434 utterances) or the interviewer (27,254 utterances). We then manually extract results sections from each research paper. In the case of mixed-methods studies, we limit our results to those that came from the qualitative interviews. We partition each results section into segments that represent the different findings from the paper.

Using the interviews, research papers, and supplemental documents like data narratives and interview plans, we add two pieces of additional data. First, we write a 1--2 sentence summary that briefly explains the overarching goals and context of the project. Then, we identify 3--5 key research questions that the study was trying to answer. Because we use the summary and research questions to analyze the characteristics of responses, as described in \sref{sec:conceptual-measures}, we ensure that \textit{they do not contain information from the final results of the paper}. Instead we align them with the initial goals of the project, as described in the interviews, research paper, and supplemental documents.

\paragraph{Excerpts} Because qualitative interviews are dialogues, they often contain overlapping speech. For example, an interviewer may say, ``mmmm'' or ``yes'' in the middle of a participant response to encourage them to continue speaking. To differentiate between a continuing participant response and a new participant response, we combine utterances into sets of excerpts. The first excerpt for each interview begins with the beginning of the transcript. Then, new excerpts are determined based on when the interviewer says more than four words. We construct 16,940 excerpts, where each excerpt begins with an interviewer utterance (most commonly a question) and contains a full participant response, occasionally interrupted by short interviewer utterances.

\section{Results}

\subsection{Can Automated Measures Capture Response Characteristics?}

\paragraph{Annotator Agreement} To validate whether our LLM judges can accurately estimate the conceptual measures, we compare their outputs to 2,400 human judgments over 100 interview excerpts. First, we compare Krippendorff's alpha between human annotators. Then, we take the median of the human labels for each excerpt and compare it to the LLM judge label using Krippendorff's alpha. We run these comparisons using both GPT-5.4 and Gemini 3.1 pro with their default settings.

\paragraph{Findings} Across the conceptual measures, we find strong agreement between human ratings and equally strong agreement between the median human ratings and the LLM judge ratings (\tref{tab:agreement_results}). GPT-5.4 shows generally stronger alignment with human ratings with an average Krippendorff's alpha of $0.74$ compared to $0.68$ for Gemini 3.1 Pro. These results show that we can automatically measure interview response characteristics and our quality criterion at scale using our LLM judge setup. This finding supports the validity of our findings in \sref{sec:regression-analysis} and enables future applications of our measures, including evaluating interview systems and informing qualitative methodology. We include a more detailed analysis of LLM judge errors in \Cref{sec:error-analysis}.

\paragraph{Prompt Sensitivity} To test the sensitivity of the LLM judges to prompt design, we rerun our comparison using GPT-5.4 with two AI-generated variations of each prompt. We find that both prompt-variations also demonstrate strong agreement with median human ratings with average Krippendorff's alphas of $0.71$ an $0.67$, though not as strong as the original prompt design. Full results are provided in \tref{tab:prompt-variations}.

\subsection{What Characteristics Are Predictive of Response Quality?}
\label{sec:regression-analysis}
To understand what makes a high-quality interview response, we evaluate which characteristics of participant responses are predictive of the response's contribution to the study findings, as measured with our response quality criterion. We use the LLM judge labels from GPT-5.4 because it has the strongest alignment with median human ratings.

\paragraph{Mixed-Effects Model} 
Because our data has a nested structure where multiple responses come from a single participant and multiple participants come from a single research project, we cannot assume independence between responses. To account for this, we use a linear mixed-effects model where the outcome is our response quality criterion, the fixed effects are the response characteristics, and the random effects are the participant and the project that the response originates from. The full equation is provided in \Cref{sec:mixed-model-equation}. Our model has a marginal $R^2$ of $0.506$, indicating that $50.6\%$ of the variation in response quality is explained by the characteristics we identify. We additionally find low multicollinearity and variance inflation factors, which support the reliability and interpretability of our model (details in \Cref{sec:multicollinearity}). We use Benjamini-Hochberg correction to adjust the p-values for simultaneous hypothesis testing.

\begin{table}[t]
\small
\centering
\begin{tabular}{@{} l c c @{}}
\toprule
\textbf{Characteristic} & \textbf{Std. Coef.} & \textbf{Std. CI} \\ \midrule
Research Question Relevance & \cellcolor{color1}\textbf{0.536} & \textbf{0.523 -- 0.549} \\ \midrule
Attributed Meaning & \cellcolor{color2}\textbf{0.137} & \textbf{0.124 -- 0.150} \\ \midrule
Specificity  & \cellcolor{color4}\textbf{0.056} & \textbf{0.041 -- 0.071} \\ \midrule
Response Length & \cellcolor{color6}\textbf{0.048} & \textbf{0.032 -- 0.064} \\ \midrule
Immediate Relevance & \cellcolor{color5}\textbf{0.039} & \textbf{0.027 -- 0.051} \\ \midrule
Spontaneity & \cellcolor{color3}\textbf{0.037} & \textbf{0.023 -- 0.051} \\ \midrule
Self-reportedness & \cellcolor{color7}\textbf{0.016} & \textbf{0.002 -- 0.029} \\  \midrule
Average Surprisal & \cellcolor{color8}0.009 & -0.002 -- 0.020 \\  \midrule
Response Length Ratio & \cellcolor{color8}0.002 & -0.012 -- 0.016 \\ \midrule
Clarity & \cellcolor{color9}0.001 & -0.011 -- 0.012 \\\bottomrule
\end{tabular}
\caption{Using a linear mixed-effect model, we identify which characteristics of participant responses are most predictive of response quality. Predictors with $p<0.05$ are in bold. The model has marginal $R^2=0.506$ and conditional $R^2=0.583$.}
\label{tab:regression}
\end{table}

\paragraph{Findings} We find that research question relevance, attributed meaning, spontaneity, specificity, immediate relevance, response length, and self-reportedness are significantly predictive of response quality (\tref{tab:regression}). Of these characteristics, research question relevance has the strongest correlation with a standard coefficient more than 3 times larger than any other covariate. This finding suggests that \textbf{the most important characteristic of high-quality responses is direct relation to a key research question} of the study.

The second strongest coefficient is for attributed meaning. Attributed meaning indicates that a response demonstrates significance or meaning to a participant. These characteristics represent a unique strength of qualitative methods that give researchers access to participants' lived experiences. Together, these two attributes demonstrate that responses are most valuable when they contribute to the overarching goals of qualitative research: answering research questions through personal insights that quantitative methods cannot capture.

Notably, we do not find statistically significant correlations for clarity, average surprisal, and response length ratio. This finding \textbf{contradicts current practice for NLP interview systems that frequently evaluate with measures of clarity and surprisal-based informativeness}. This finding also aligns with common advice in qualitative literature that participants speaking clearly or having an extensive vocabulary does not guarantee that they are best suited to provide valuable insights. Note that using total surprisal instead of average surprisal and response length results in a standard coefficient of $0.041$ without changing the marginal $R^2$ or the other standard coefficients. This indicates that the surprisal measure itself does not provide predictive power beyond being a proxy for response length. Full regression results are provided in \tref{tab:full-regression}.

\paragraph{Robustness} To test whether these the results of our mixed-effect regression are robust to model and prompt selection, we relabel the full dataset with three variations of the LLM judges. We use Gemini 3.1 Pro with the original prompts and GPT-5.4 with two AI-generated prompt variations. Using the labels from the LLM judge variations, we rerun the above regression and find that our results are consistent. In all runs, research question relevance is by far the most important predictor. There is some variation in the smaller coefficient predictors, but in 3 of 4 runs clarity is statistically insignificant, and in 3 of 4 runs average surprisal is statistically insignificant. These results demonstrate the robustness of our findings across differences in model and prompt selection. Full details are provided in \Cref{sec:prompt-and-model-selection}.

We additionally consider whether the insignificance of clarity and surprisal could be due to attenuation bias caused by measurement error in the LLM  judges. To test this, we run a simulation extrapolation (SIMEX) procedure that injects noise into our measures according to their reliability as measured by Krippendorff's alpha. Again, we find that these results are consistent with the original regression. Research question relevance is still the most important predictor, and clarity and average surprisal are still statistically insignificant. Full details are provided in \Cref{sec:attenuation-bias}.

\subsection{Case Study: How Do Techniques and Time Affect Quality?}
\label{sec:techniques-analysis}

\begin{figure}
    \centering
    \includegraphics[width=1\linewidth]{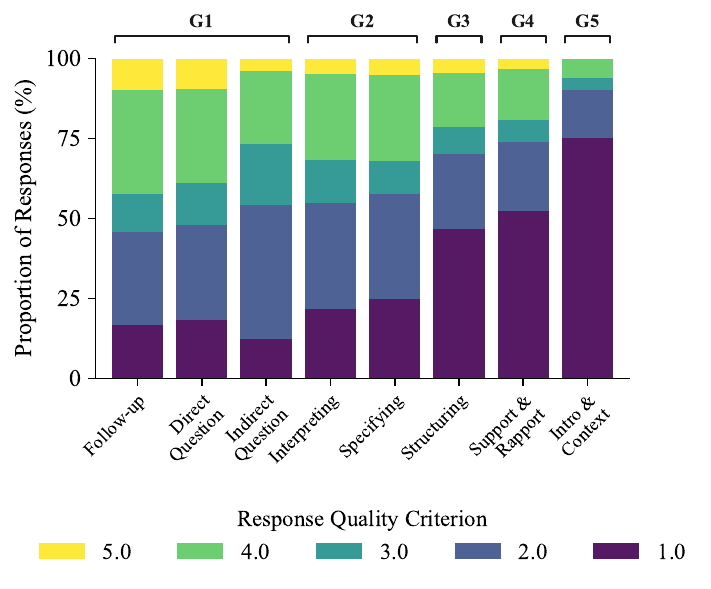}
    \caption{We compare the distributions of quality in responses elicited using different interview techniques. We group techniques into five distinct groups (G1--G5) that correspond to the theoretical function of the techniques in the group. Each member of a group has a statistically significant difference in median with the members of all other groups.}
    \label{fig:techniques}
\end{figure}

\begin{figure}
    \centering
    \includegraphics[width=1\linewidth]{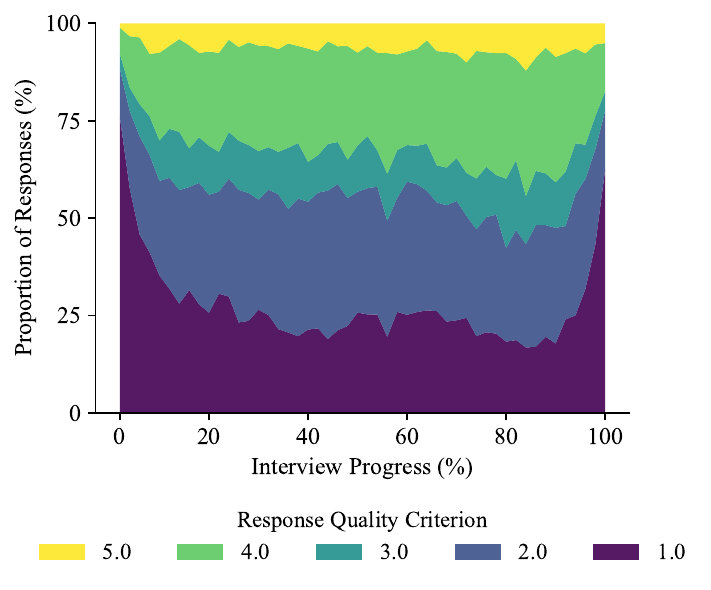}
    \caption{We compare the distribution of quality responses over the course of the interviews. Our quality criterion reveals a temporal trend where participants tend to provide the lowest quality responses during the beginnings and ends of interviews.}
    \label{fig:quality_over_time}
\end{figure}

The ultimate goal of assessing response quality is to inform decisions about interview strategies and interviewer system design. To highlight the potential for our methods to inform those decisions, we conduct a case study of how interview techniques affect response quality and how response quality changes over time.

\paragraph{Interview Techniques} Using a similar LLM judge setup as \sref{sec:conceptual-measures}, we prompt the model to identify relevant techniques that the interviewer used in an excerpt based on \citeposs{kvale_interviews_2009}'s taxonomy of interview techniques (\tref{tab:techniques}). We use the same annotation setup as before to validate these judgments by comparing them to 300 human judgments over 100 excerpts. Because excerpts can contain multiple techniques, we compare the average Jaccard similarity. We find that the similarity between pairs of human annotators is $0.51$, compared to $0.5$ between the LLM judge and the human annotators.

We classify excerpts based on the interview techniques used in them and then use a Kruskal-Wallis test to find a statistically significant difference in median response quality for responses obtained with the different techniques ($p<0.001$). We then use Dunn's post-hoc test with Bonferroni correction to identify statistically significant differences in medians between pairs and use those to identify groups of techniques with similar response quality. Full details of the test are provided in \Cref{sec:dunns-test}.

\fref{fig:techniques} displays the results. Each member of a group has a statistically significant difference in median response quality as compared to members of all other groups. We find that techniques that are used to elicit information core to the research project (Group 1: Follow-up, Direct Questioning, Indirect Questioning) result in interview responses with the highest quality ratings. The group with the second-highest quality responses represents techniques that are designed to clarify responses or reach common ground with a participant (Group 2: Specifying, Interpreting). The group with the third-highest quality responses are techniques that guide or direct the attention of participants (Group 3: Structuring). The group with the fourth-highest quality responses aims to build rapport with the participant to elicit higher quality responses later in the interview (Group 4: Support \& Rapport Building). The final group represents techniques that explain the project to the participant and collect background information (Group 5: Introduction \& Contextualization).

\paragraph{Time} We also analyze the effects of time on the quality of responses. Because our dataset has interviews of varying lengths, we normalize the time as the progress through the total length of the interview from 0 to 100\% and compare the distribution of response quality over the normalized time (\fref{fig:quality_over_time}). Our results reveal a temporal trend where participants tend to provide the lowest quality responses during the beginnings and ends of interviews. This finding is consistent with common interview timelines where interviewers reserve the beginnings and ends of interviews for logistics, small talk, and winding down.

These results demonstrate that our measures capture meaningful differences in response quality that reflect both the different functions of various interview techniques and common timelines of interviews.
Future work could use our measures to investigate how interview techniques affect interview quality more deeply, such as if conducting rapport building and contextualization early in an interview improves later responses to direct questions.

\section{Discussion}
Our results provide insights to inform designers of NLP interview systems and qualitative researchers.

\textit{For interview system designers}, we provide empirically grounded metrics to evaluate the quality of the data that NLP interview systems collect. Designers should evaluate relevance to a research question, as it is the most predictive of contribution to a study's findings.

In contrast, researchers should not emphasize clarity and surprisal-based informativeness, as we do not find evidence that they are predictors of contribution to a study's findings.
We further show that all the metrics we investigate can be measured at scale with LLMs, thus facilitating automated evaluation. These metrics could also be used as a reinforcement learning objective to train an interview system.

\textit{For qualitative researchers}, we validate theoretical frameworks of response quality that can be used to guide qualitative studies (e.g., helping researchers identify when they need to modify interview plans), to train qualitative researchers, and to conduct new studies, like evaluating the effects of different interview techniques or participant selection methods on response quality.

Future work can build on our methods by going beyond individual responses and analyzing quality across a full interview context to capture the effects of time-dependent techniques like rapport building. They can also explore connections between our work and data saturation to better quantify not just whether responses contribute to the results, but how they contribute in comparison to one another.

\section{Related Work}

\subsection{Empirical Analysis of Qualitative Interviews}

Limited prior work has empirically analyzed qualitative interviews. Surveys on qualitative methodology have been conducted to understand perspectives and common practices \citep{muthanna_interrelationship_2023, salet_good_2025}, but these focus on the methodology rather than the data that is collected.

One barrier to this type of empirical analysis is the availability of qualitative data. To our knowledge, there is no openly available dataset for analyzing qualitative interview projects across multiple domains. We address this gap by introducing the Qualitative Interview Corpus, a dataset of 343 qualitative interviews across a diverse set of domains that will enable further empirical analysis of the data and results from qualitative research projects.

\subsection{Data Quality in Qualitative Research}

Discussions of quality in qualitative literature primarily focus on methodological rigor rather than evaluating the collected data \citep{tracy_qualitative_2010, roulston_considering_2010, cope_methods_2014, korstjens_series_2018} because there is an assumption that a human investigator is directing the research project towards its objectives. However, there are cases that require evaluating the quality of the data itself. For example, experimenting with new interview strategies or evaluating NLP interview systems, which are capable of conducting methodologically rigorous interviews that do not collect any useful data for the project's goals.

As detailed in \sref{sec:proposed-characteristics}, some work in qualitative research has proposed characteristics of high-quality interview responses \citep{kvale_interviews_2009, charmaz_constructing_2014, patton_qualitative_2015, small_qualitative_2022}, but these frameworks are based on personal experience and lack empirical evidence to validate them, leading to a lack of clarity in evaluating interview response quality. Our work addresses this gap by empirically evaluating the extent to which responses with these characteristics actually contribute to the results of the study.

\subsection{Evaluating Interview Systems}
The lack of clarity in measuring interview data quality has translated to a lack of clarity in evaluating NLP interview systems. Some system objectives, like engaging participants \citep{xiao_tell_2020, cuevas_collecting_2025} or maintaining coherent conversation \citep{guo_pcqpr_2024, liu_mimitalk_2025} have intuitive measures, but there is a lack of empirically grounded measures for data quality.

Work in information elicitation has attempted to measure response quality by explicitly modeling belief distributions and information gain \citep{handa_bayesian_2024, choudhury_bed-llm_2025}. However, these methods are not appropriate for most qualitative interviews, where investigators aim to collect nuanced insights that cannot be clearly mapped onto a probability distribution. 

Other work has measured quality with domain expert judgments of the insights revealed in the interviews \citep{anugraha_sparkme_2026}. Though this is a robust method for measuring the final output of an interview system, it is expensive and impractical for many important tasks like intermediate evaluations, defining an objective function, or comparing large numbers of systems.

The most popular method for measuring response quality is using other characteristics of participant responses, such as specificity, clarity, and relevance, as proxy criteria. These characteristics are chosen with theoretical justifications coming from the Gricean maxims \citep{xiao_if_2020, xiao_tell_2020, jiang_communitybots_2023, hu_designing_2024, jacobsen_chatbots_2025} or qualitative literature \citep{cuevas_collecting_2025}. Our work provides the empirical validation missing in prior studies, showing which metrics actually translate to achieving the goals of the study.

\section{Conclusion}
In this work, we introduce the Qualitative Interview Corpus and use it to empirically evaluate which proposed measures of interview response quality are actually predictive of a response's contribution to the study findings. We find that the strongest predictor of response quality is relevance to a key research question, and we show that two commonly used metrics, clarity and surprisal-based informativeness, are not predictive of response quality. Our work highlights the importance of empirically validating theoretical frameworks in qualitative research and enables future research to understand qualitative interviews and evaluate interview systems.

\section*{Limitations}
The primary limitation of our work is that our analysis is conducted over 14 qualitative interview projects. While we collect projects that cover a range of topics and study populations, as described in \Cref{sec:descriptive-stats}, we cannot be certain that our results would generalize to new studies. To mitigate this limitation, we describe our framework in detail and release our code to support running our evaluation on other studies.

Our work additionally focuses on the perspective of the researcher and interviewer, in that our assessment of interview quality is focused on what aspects of the interview contributed to the final results of the paper. They do not capture the interviewee's perspective, such as whether or not the interviewee felt comfortable and engaged.

Finally, we treat inclusion in paper results as a ``ground truth'' metric of interview quality, which assumes that researchers correctly analyzed interview content. In practice, qualitative researchers may have missed relevant content provided by the interviewee.

\section*{Ethical Considerations}
We have coordinated with the Qualitative Data Repository to ensure our use of this data is within the terms of service of their platform and abides by the user agreements that researchers agreed to when uploading data to the platform. The Qualitative Data Repository now houses our processed data on their platform under the same terms of use as the original unprocessed transcripts. As our work constitutes secondary analysis of publicly available de-identified data collected for research purposes, there are no risks that we know of to study participants or researchers included in this data.

\section*{Acknowledgments}
We thank the Qualitative Data Repository for hosting the Qualitative Interview Corpus and working with us to ensure proper use of their data. We also thank Mateus Mendonça, Didi Zhou, Yuka Iijima, and Jennifer Leano for their help.

\bibliography{custom}

\appendix

\clearpage
\raggedbottom
\section{Model Prompts}
\label{sec:model_prompts}

We provide the exact prompts used in the LLM judges for the response characteristics, quality criterion, and interview techniques (Figures~\ref{fig:prompt_attributed_meaning}--\ref{fig:prompt_interviewer_techniques}).

\begin{figure*}[ht]
\begin{tcolorbox}[
    colback=gray!5!white, 
    colframe=gray!50!black, 
    title=Prompt: Attributed Meaning
]
You are an expert qualitative researcher analyzing interview data.

Rate the level of attributed meaning of the participant statement from in the current interview excerpt below on a scale from 1 to 3. In addition to the current excerpt, you are also provided with a short context blurb and the interview excerpt that immediately preceded the current excerpt in the transcript. These two sections are only to understand the context of the current excerpt, and your rating should be for participant statement in the current excerpt.

Scoring Rubric:\\
1. The participant statement does not provide information indicating how significant an action or experience is to the participant.\\
2. The participant statement indicates some level of significance of an action or experience to a participant but does not provide information about what that significance is.\\
3. The participant statement directly shows the significance of an action or experience to the participant and explains what that significance is.

CONTEXT BLURB (context only):\\
\{context\_blurb\}

PREVIOUS INTERVIEW EXCERPT (context\_only):\\
\{previous\}

CURRENT INTERVIEW EXCERPT (rate this):\\
\{excerpt\}

CRITICAL: Output only a single digit (1, 2, or 3). Do not write any additional text.
\end{tcolorbox}
\caption{LLM Judge prompt used to evaluate the attributed meaning of participant statements.}
\label{fig:prompt_attributed_meaning}
\end{figure*}

\begin{figure*}[ht]
\begin{tcolorbox}[
    colback=gray!5!white, 
    colframe=gray!50!black, 
    title=Prompt: Clarity
]
You are an expert qualitative researcher analyzing interview data.

Rate the clarity of the participant statement from in the current interview excerpt below on a scale from 1 to 3. In addition to the current excerpt, you are also provided with a short context blurb and the interview excerpt that immediately preceded the current excerpt in the transcript. These two sections are only to understand the context of the current excerpt, and your rating should be for participant statement in the current excerpt.

Scoring Rubric:\\
1. The participant statement is incoherent, or the meaning is completely unclear.\\
2. The participant statement is ambiguous or requires guessing to understand.\\
3. The participant statement is clear to read and understand.

CONTEXT BLURB (context only):\\
\{context\_blurb\}

PREVIOUS INTERVIEW EXCERPT (context\_only):\\
\{previous\}

CURRENT INTERVIEW EXCERPT (rate this):\\
\{excerpt\}

CRITICAL: Output only a single digit (1, 2, or 3). Do not write any additional text.
\end{tcolorbox}
\caption{LLM Judge prompt used to evaluate the clarity of participant statements.}
\label{fig:prompt_clarity}
\end{figure*}

\begin{figure*}[ht]
\begin{tcolorbox}[
    colback=gray!5!white, 
    colframe=gray!50!black, 
    title=Prompt: Immediate Relevance
]
You are an expert qualitative researcher analyzing interview data.

Rate how relevant the participant's statement is to the specific question asked by the interviewer in the current interview excerpt below on a scale from 1 to 3. In addition to the current excerpt, you are also provided with a short context blurb and the interview excerpt that immediately preceded the current excerpt in the transcript. These two sections are only to understand the context of the current excerpt, and your rating should be for participant statement in the current excerpt.

Scoring Rubric:\\
1. The participant statement is completely unrelated to the question asked, avoids the question entirely, or addresses a totally different topic.\\
2. The participant statement is related to the general topic of the question but drifts or answers a different question than the one posed.\\
3. The participant statement directly answers the specific question posed by the interviewer.

CONTEXT BLURB (context only):\\
\{context\_blurb\}

PREVIOUS INTERVIEW EXCERPT (context\_only):\\
\{previous\}

CURRENT INTERVIEW EXCERPT (rate this):\\
\{excerpt\}

CRITICAL: Output only a single digit (1, 2, or 3). Do not write any additional text.
\end{tcolorbox}
\caption{LLM Judge prompt used to evaluate the relevance of participant statements to the interviewer's question.}
\label{fig:prompt_relevance_interviewer}
\end{figure*}
\begin{figure*}[ht]
\begin{tcolorbox}[
    colback=gray!5!white, 
    colframe=gray!50!black, 
    title=Prompt: Self-Reportedness
]
You are an expert qualitative researcher analyzing interview data.

Rate the self-reportedness of the participant statement from in the current interview excerpt below on a scale from 1 to 3. In addition to the current excerpt, you are also provided with a short context blurb and the interview excerpt that immediately preceded the current excerpt in the transcript. These two sections are only to understand the context of the current excerpt, and your rating should be for participant statement in the current excerpt.

Scoring Rubric:\\
1. The participant statement is not interpretable without additional context.\\
2. The core idea of the participant statement is understandable but may require additional context for full understanding.\\
3. The participant statement is fully self-contained, not requiring any additional context to be interpretable.

CONTEXT BLURB (context only):\\
\{context\_blurb\}

PREVIOUS INTERVIEW EXCERPT (context\_only):\\
\{previous\}

CURRENT INTERVIEW EXCERPT (rate this):\\
\{excerpt\}

CRITICAL: Output only a single digit (1, 2, or 3). Do not write any additional text.
\end{tcolorbox}
\caption{LLM Judge prompt used to evaluate the self-reportedness of participant statements.}
\label{fig:prompt_self_reportedness}
\end{figure*}

\begin{figure*}[ht]
\begin{tcolorbox}[
    colback=gray!5!white, 
    colframe=gray!50!black, 
    title=Prompt: Specificity
]
You are an expert qualitative researcher analyzing interview data.

Rate the specificity of the participant statement from in the current interview excerpt below on a scale from 1 to 3. In addition to the current excerpt, you are also provided with a short context blurb and the interview excerpt that immediately preceded the current excerpt in the transcript. These two sections are only to understand the context of the current excerpt, and your rating should be for participant statement in the current excerpt.

Scoring Rubric:\\
1. The participant statement is generic or abstract, providing only high-level summaries or vague descriptions.\\
2. The participant statement describes a particular event, action, or opinion without concrete details or examples.\\
3. The participant statement describes a particular event, action, or opinion with concrete details or examples.

CONTEXT BLURB (context only):\\
\{context\_blurb\}

PREVIOUS INTERVIEW EXCERPT (context\_only):\\
\{previous\}

CURRENT INTERVIEW EXCERPT (rate this):\\
\{excerpt\}

CRITICAL: Output only a single digit (1, 2, or 3). Do not write any additional text.
\end{tcolorbox}
\caption{LLM Judge prompt used to evaluate the specificity of participant statements.}
\label{fig:prompt_specificity}
\end{figure*}

\begin{figure*}[ht]
\begin{tcolorbox}[
    colback=gray!5!white, 
    colframe=gray!50!black, 
    title=Prompt: Spontaneity
]
You are an expert qualitative researcher analyzing interview data.

Rate the spontaneity of the participant statement from in the current interview excerpt below on a scale from 1 to 3. In addition to the current excerpt, you are also provided with a short context blurb and the interview excerpt that immediately preceded the current excerpt in the transcript. These two sections are only to understand the context of the current excerpt, and your rating should be for participant statement in the current excerpt.

Scoring Rubric:\\
1. The participant statement only confirms or reiterates information provided in the interviewer’s statement.\\
2. The participant statement adds additional information beyond what was provided in the interviewer’s statement but remains within the topic posed.\\
3. The participant statement introduces a new topic that may be related but was not introduced in the interviewer’s statement.

CONTEXT BLURB (context only):\\
\{context\_blurb\}

PREVIOUS INTERVIEW EXCERPT (context\_only):\\
\{previous\}

CURRENT INTERVIEW EXCERPT (rate this):\\
\{excerpt\}

CRITICAL: Output only a single digit (1, 2, or 3). Do not write any additional text.
\end{tcolorbox}
\caption{LLM Judge prompt used to evaluate the spontaneity of participant statements.}
\label{fig:prompt_spontaneity}
\end{figure*}

\begin{figure*}[ht]
\begin{tcolorbox}[
    colback=gray!5!white, 
    colframe=gray!50!black, 
    title=Prompt: Research Question Relevance
]
You are an expert qualitative researcher analyzing interview data.

Estimate how relevant the participant statement from the current interview excerpt below is to the provided research question on a scale from 1 to 3. In addition to the current excerpt and research question, you are also provided with a short context blurb and the interview excerpt that immediately preceded the current excerpt in the transcript. These two sections are only to understand the context of the current excerpt, and your rating should be for participant statement in the current excerpt.

Scoring Rubric:\\
1. The participant statement is unrelated to the research question or discusses a completely different topic.\\
2. The participant statement is tangentially related to the topic of the research question.\\
3. The participant statement directly addresses the research question.

RESEARCH QUESTION:\\
\{research\_question\}

CONTEXT BLURB (context only):\\
\{context\_blurb\}

PREVIOUS INTERVIEW EXCERPT (context\_only):\\
\{previous\}

CURRENT INTERVIEW EXCERPT (rate this):\\
\{excerpt\}

CRITICAL: Output only a single digit (1, 2, or 3). Do not write any additional text.
\end{tcolorbox}
\caption{LLM Judge prompt used to evaluate the relevance of participant statements to a key research question.}
\label{fig:prompt_relevance_rq}
\end{figure*}

\begin{figure*}[ht]
\begin{tcolorbox}[
    colback=gray!5!white, 
    colframe=gray!50!black, 
    title=Prompt: Quality Criterion
]
You are an expert qualitative researcher analyzing interview data.

Your task is to rate the likelihood that the participant statement from the current interview excerpt below contributed to the provided results section on a scale from 1-5. In addition to the current excerpt, you are also provided with a short context blurb and the interview excerpt that immediately preceded the current excerpt in the transcript. These two sections are only to understand the context of the current excerpt, and your rating should be for participant statement in the current excerpt.

Scoring Rubric:\\
1. The statement is unrelated to the results section or contradicts it.\\
2. Tangential relation; discusses the topic but offers no specific substance.\\
3. Aligns with the results section but is general or vague.\\
4. Provides an example or sentiment that matches the results section’s conclusion.\\
5. Appears in the results section and likely served as a primary source for it.

RESULTS SECTION:\\
\{result\}

CONTEXT BLURB (context only):\\
\{context\_blurb\}

PREVIOUS INTERVIEW EXCERPT (context\_only):\\
\{previous\}

CURRENT INTERVIEW EXCERPT (rate this):\\
\{excerpt\}

CRITICAL: Output only a single digit (1, 2, 3, 4, or 5). Do not write any additional text.
\end{tcolorbox}
\caption{LLM Judge prompt used to evaluate the likelihood that participant statements contributed to the results section.}
\label{fig:prompt_contribution}
\end{figure*}

\begin{figure*}[ht]
\begin{tcolorbox}[
    colback=gray!5!white, 
    colframe=gray!50!black, 
    title=Prompt: Interview Techniques
]
You are an expert qualitative researcher analyzing interview data.

Your task is to analyze the interviewer statement in the current interview excerpt below and determine which of the following categories it fits in based on the strategy that the interviewer is using. In addition to the current excerpt, you are also provided with a short context blurb and the interview excerpt that immediately preceded the current excerpt in the transcript. These two sections are only to understand the context of the current excerpt, and your rating should be for interviewer statement in the current excerpt.

Possible Categories:\\
1. Introductory/Contextualization Questions: Open-ended questions that may be unrelated to the overall research questions but are designed to give an understanding of the participant or context.\\
2. Support and Rapport Building: A statement designed to make a connection with the participant, provide support, or let the participant know that the purpose of the interview is being fulfilled.\\
3. Follow-up / Elaboration Probe: A statement designed to encourage a participant to continue talking. It may be a simple “uh-huh” or “mmm,” or it could also be a direct call such as, “Could you say some more about that?”\\
4. Specifying / Detail-Oriented Probe: Questions that follow up to ask who, where, what, when, or how to obtain a complete and detailed picture of an activity or experience.\\
5. Direct Questioning: A question that directly introduces topics or dimensions and asks the respondent about them.\\
6. Indirect / Projective Questioning: Indirect questions that may ask about the attitudes of others or encourage an indirect statement of the participant’s own motivations, attitudes, or emotions.\\
7. Structuring: A statement that controls the structure of the interview by transitioning topics, redirecting respondents, or breaking off participant answers that may be irrelevant to the purpose of the interview.\\
8. Interpreting: A statement that rephrases or interprets answers provided by the participant to get clarification or reach common ground with the participant.\\

CONTEXT BLURB (context only):\\
\{context\_blurb\}

PREVIOUS INTERVIEW EXCERPT (context\_only):\\
\{previous\}

CURRENT INTERVIEW EXCERPT (rate this):\\
\{current\_excerpt\}

CRITICAL: Output only digits (1, 2, 3, 4, 5, 6, 7, 8) separated by commas. Do not write any additional text.
\end{tcolorbox}
\caption{LLM Judge prompt used to identify the techniques used in interviewer statements.}
\label{fig:prompt_interviewer_techniques}
\end{figure*}
\clearpage

\begin{table*}[th!]
\small
\centering
\begin{tabular*}{\textwidth}{@{\extracolsep{\fill}} l c c c c @{} }
\toprule
\textbf{Metric} & \textbf{Word Count} & \textbf{Interviewer Utterances} & \textbf{Participant Utterances} & \textbf{Excerpts} \\ \midrule
Total & 2,157,939 & 27,254 & 31,434 & 16,940 \\ \midrule
Average Per Project & 154,138.50 & 1,946.71 & 2,245.29 & 1,210 \\ \midrule
Average Per Interview & 6,147.97 & 79.46 & 91.64 & 49.39 \\ \midrule
Average Per Excerpt & 127.39 & 1.61 & 1.86 & — \\ \bottomrule
\end{tabular*}
\caption{The word count, number of interviewer utterances, number participant utterances, and number of excerpts in the Qualitative Interview Corpus.}
\label{tab:descriptive-stats}
\end{table*}

\section{Qualitative Interview Corpus Construction}

\subsection{Annotation Setup}
\label{sec:annotation_setup}

\begin{figure*}[h]
    \centering
    \includegraphics[width=1\linewidth]{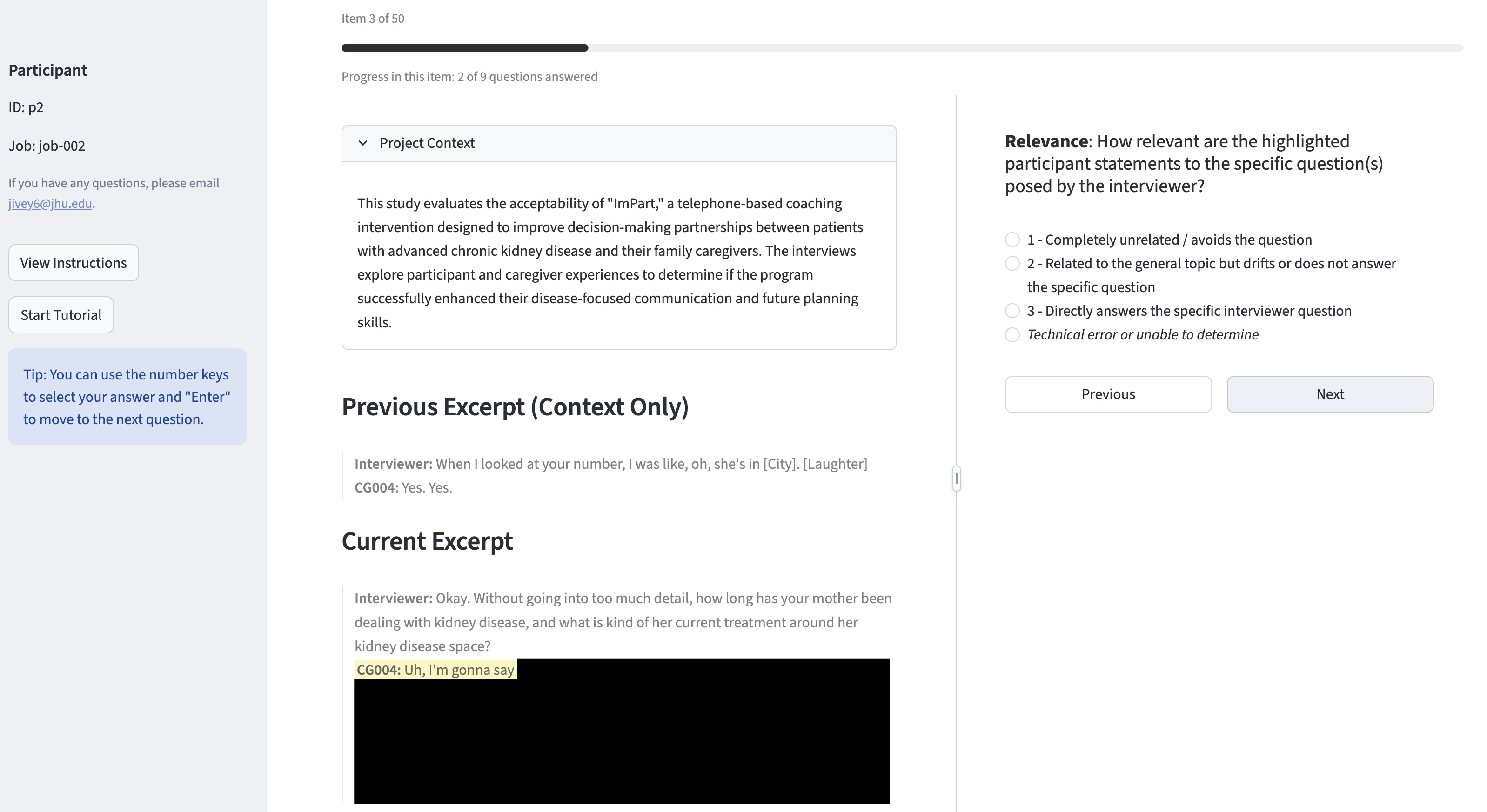}
    \caption{An example of the interface used to collect human annotations. The participant response is redacted to comply with the Qualitative Data Repository's terms of use.}
    \label{fig:annotation-setup}
\end{figure*}

To validate whether LLM judgments can be used to operationalize our conceptual measures, we designed an annotation task, as displayed in \fref{fig:annotation-setup}. We recruited five graduate students with experience analyzing qualitative interviews to rate either 50 or 100 excerpts and compensated them at \$20 per hour.

\subsection{Details of the Qualitative Interview Corpus}
\label{sec:descriptive-stats}

The Qualitative Interview Corpus is composed of 343 qualitative interviews and their corresponding papers from 14 research projects across a diverse set of domains, including healthcare and emergency work \citep{F6AGWUJG_2021, F68TOJJY_2024, F6FYZITI_2025}, political engagement and representation \citep{F6HYTYIJ_2025, F6L9HHYL_2025}, nutrition \citep{F6MTPVK7_2025}, defense \citep{F6AHDRFQ_2020}, governance and organization \citep{F6QHVGUI_2023}, spiritual care \citep{F6R7J9HL_2025}, palliative care \citep{F6YMWPUX_2021, F6UXQABW_2024}, education \citep{F6XZV8BZ_2025}, labor \citep{F6Z82KER_2024}, and youth athletics \citep{F6ZP448B_2017}. The projects primarily use a semi-structured interview format, with some including additional components such as structured classroom observation \citep{F6XZV8BZ_2025}, policy analysis \citep{F6QHVGUI_2023}, user testing \citep{F6R7J9HL_2025}, and cross-case comparative designs \citep{F6Z82KER_2024}. Four of the projects used mixed-methods designs that combine qualitative and quantitative approaches \citep{F6QHVGUI_2023, F6Z82KER_2024, F6HYTYIJ_2025, F6L9HHYL_2025}. The projects are skewed towards U.S. geography with only four other countries represented: Ghana \citep{F6FYZITI_2025}, Ethiopia \citep{F6MTPVK7_2025}, South Africa \citep{F6L9HHYL_2025}, and Australia \citep{F6ZP448B_2017}. Within the U.S., the studies have broad coverage of the major geographic regions. Additional details about the projects used are provided in \tref{tab:project-details}, and details about the composition of the projects, interviews, and excerpts are provided in \tref{tab:descriptive-stats}.

\begin{table}[t]
\centering
\small
\begin{tabular}{p{5.5cm} r} %
\toprule
\textbf{Predictor Variable} & \textbf{VIF} \\ \midrule
Response Length & 2.25 \\
Specificity & 2.21 \\
Spontaneity & 1.86 \\
Attributed Meaning & 1.75 \\
Self-reportedness & 1.69 \\
Response Length Ratio & 1.64 \\
Research Question Relevance & 1.60 \\
Immediate Relevance & 1.39 \\
Clarity & 1.25 \\
Average Surprisal & 1.06 \\ \bottomrule
\end{tabular}
\caption{Variance Inflation Factors (VIF) for our mixed-effects model.}
\label{tab:variance-inflation}
\end{table}

\begin{table}[t]
\small
\centering
\begin{tabular}{l c c}
\toprule
\textbf{Characteristic} & \textbf{Raw Coef.} & \textbf{P-Value} \\ \midrule
\shortstack[l]{Research Question Relevance} & \textbf{1.320} & \textbf{<0.001} \\ \midrule
Attributed Meaning           & \textbf{0.210} & \textbf{<0.001} \\ \midrule
Specificity                  & \textbf{0.080} & \textbf{0.002}  \\ \midrule
Response Length              & \textbf{0.000} & \textbf{0.016}  \\ \midrule
Immediate Relevance          & 0.040          & 0.064          \\ \midrule
Spontaneity                  & 0.002          & 0.958          \\ \midrule
Self-reportedness            & -0.013         & 0.548          \\ \midrule
Average Surprisal            & 0.017          & 0.123          \\ \midrule
Response Length Ratio        & 0.000          & 0.939          \\ \midrule
Clarity                      & 0.011          & 0.719          \\ \bottomrule
\end{tabular}
\caption{An error-aware regression using SIMEX estimation finds results consistent with our original mixed-effect model, where research question relevance is the most important predictor and clarity and average surprisal are not statistically significant. Bold indicates p-values $<0.05$.}
\label{tab:simex}
\end{table}

\begin{table}[t]
\small
\centering
\begin{tabular}{@{} l p{5cm} @{}}
\toprule
\textbf{Technique} & \textbf{Description} \\ \midrule
\makecell[tl]{Introduction \& \\ Contextualization} & Open-ended questions designed to understand the participant or context, often unrelated to core research questions. \\ \midrule
\makecell[tl]{Support \& \\ Rapport Building} & Statements designed to build a connection, provide support, or validate the participant's contribution. \\ \midrule
\makecell[tl]{Follow-up} & Brief interjections (e.g., "uh-huh") or direct calls to encourage the participant to continue talking. \\ \midrule
\makecell[tl]{Specifying} & Follow-up questions (who, what, where, when, how) to obtain a detailed picture of an experience. \\ \midrule
\makecell[tl]{Direction \\ Questioning}  & Questions that directly introduce specific topics or dimensions to the respondent. \\ \midrule
\makecell[tl]{Indirect \\ Questioning} & Questions about others' attitudes to indirectly surface the participant's own motivations or emotions. \\ \midrule
Structuring & Statements used to transition topics, redirect respondents, or interrupt irrelevant answers. \\ \midrule
Interpreting & Rephrasing or interpreting answers to seek clarification or reach common ground. \\ \bottomrule
\end{tabular}
\caption{Taxonomy of interview techniques from \citet{kvale_interviews_2009}}
\label{tab:techniques}
\end{table}

\begin{table*}[h]
\footnotesize
\centering
\begin{tabularx}{\textwidth}{@{} >{\raggedright\arraybackslash}p{3.8cm} >{\raggedright\arraybackslash}p{2.5cm} >{\raggedright\arraybackslash}X c >{\centering\arraybackslash}p{1.5cm} @{}}
\toprule
\textbf{Research Project} & \textbf{Subjects} & \textbf{Keywords} & \textbf{\# Interviews} & \textbf{Avg. Word Count} \\ \midrule

Mindfulness for Firefighters and EMS Workers \citep{F68TOJJY_2024} & Medicine, Health and Life Sciences & firefighters, mindfulness, emergency medical service (EMS) providers, barriers, facilitators, implementation & 11 & 5,520 \\ \midrule

Drug Shortage Management \citep{F6AGWUJG_2021} & Medicine, Health and Life Sciences & pharmacy, inventory control, inventory shortages, cooperation, drug shortages & 16 & 5,462 \\ \midrule

Ghanaian Healthcare Workers During COVID-19 \citep{F6FYZITI_2025} & Medicine, Health and Life Sciences & COVID-19, healthcare worker & 20 & 3,891 \\ \midrule

Socializing Policy Feedback \citep{F6HYTYIJ_2025} & Social Sciences & adolescence, welfare, politics, attitude, civic, government, youth, policy & 30 & 6,039 \\ \midrule

Perspectives on Political Representation \citep{F6L9HHYL_2025} & Social Sciences & political representation, politics, voting & 23 & 2,588 \\ \midrule

Nutrition Interventions in Rural Ethiopia \citep{F6MTPVK7_2025} & Medicine, Health and Life Sciences & nutrition, nutrition-sensitive, nutrition-specific, community health, agriculture, multisectoral & 21 & 1,535 \\ \midrule

Marine Corps Education Project \citep{F6AHDRFQ_2020} & Medicine, Health and Life Sciences; Social Sciences & stress, resilience, training and education, organizational values, biological determination, armed forces, applied social science, combat stress & 32 & 7,527 \\ \midrule

Intergovernmental Coordination Mechanisms \citep{F6QHVGUI_2023} & Earth and Environmental Sciences; Social Sciences & coordination, groundwater, sustainability, inter-organizational relationships, water utilities & 43 & 9,198\\ \midrule

Models of Delivery for Online Spiritual Care \citep{F6R7J9HL_2025} & Computer and Information Science & spiritual care, chaplaincy, healthcare, nursing, palliative care, mental health, religion, spirituality & 21 & 12,123 \\ \midrule

Partnership between Kidney Disease Patients and Caregivers \citep{F6UXQABW_2024} & Medicine, Health and Life Sciences & decision making, training, program evaluation, chronic illnesses, renal disease, healthcare delivery & 25 & 2,647 \\ \midrule

Shared Data for Learning Qualitative Data Analysis \citep{F6XZV8BZ_2025} & Social Sciences & active learning, teaching methods, college students, college faculty, qualitative research & 9 & 6,082 \\ \midrule

Advance Care Planning in Hospice Organizations \citep{F6YMWPUX_2021} & Medicine, Health and Life Sciences; Social Sciences & hospices, life care planning, palliative treatment, goals of care & 50 & 6,828 \\ \midrule

Food Retail and Service Workers during COVID-19 \citep{F6Z82KER_2024} & Medicine, Health and Life Sciences; Social Sciences & precarious employment, employment quality, fundamental causes, constrained choices, policy, COVID-19 & 23 & 11,369 \\ \midrule

High-performance School-age Athletes at Australian Schools \citep{F6ZP448B_2017} & Social Sciences & athlete, bullying, high performance, NVivo, parent, school age, schools, student-athlete, teacher & 19 & 2,021\\ \bottomrule

\end{tabularx}
\caption{A detailed view of the research projects used in the Qualitative Interview Corpus including their self-identified subjects and keywords from the Qualitative Data Repository, the total number of interviews that they contributed to the corpus, and the average length of their interviews measured in number of words.}
\label{tab:project-details}
\end{table*}

\section{LLM Judges}

\begin{table*}[h]
\small
\centering
\begin{tabular*}{\textwidth}{@{\extracolsep{\fill}} l c c c c @{} } %
\toprule
\textbf{Conceptual Measure} & \textbf{\makecell{Human\\Agreement}} & \textbf{\makecell{GPT-Human \\ (Original)}} & \textbf{\makecell{GPT-Human \\ (Variation 1)}} & \textbf{\makecell{GPT-Human \\ (Variation 2)}} \\ \midrule
Attributed Meaning & 0.750 & 0.868 & 0.881 & 0.860 \\ \midrule
Spontaneity & 0.716 & 0.797 & 0.790 & 0.751 \\ \midrule
Specificity & 0.732 & 0.789 & 0.812 & 0.790 \\ \midrule
Immediate Relevance & 0.602 & 0.764 & 0.659 & 0.536 \\ \midrule
Response Quality & 0.754 & 0.757 & 0.601 & 0.604 \\ \midrule
\makecell[tl]{Research Question Relevance} & 0.714 & 0.690 & 0.633 & 0.668 \\ \midrule
Self-reportedness & 0.754 & 0.679 & 0.695 & 0.580 \\ \midrule
Clarity & 0.598 & 0.606 & 0.615 & 0.571 \\ \bottomrule
\end{tabular*}
\caption{AI-generated variations of the LLM judge prompts slightly under perform the original prompt design, but maintain strong agreement with median human ratings.}
\label{tab:prompt-variations}
\end{table*}

\subsection{Prompt Sensitivity}
\label{sec:prompt-variations}

To test the sensitivity of the LLM judges to prompt design, we rerun our comparison to median human ratings using GPT-5.4 with two AI-generated variations of each prompt (\tref{tab:prompt-variations}). Both variations still have strong agreement with the median human ratings but not as strong as the original prompt design.

\subsection{Error Analysis}
\label{sec:error-analysis}
To understand cases where LLM judgments diverge from human annotations, we analyzed the five excerpts that resulted in the highest difference in LLM and human judgment for each conceptual measure. Across all the measures we identify two commonalities in responses that LLMs mislabel: short responses and domain-specific language. In 10 out of the 40 high-disagreement cases we reviewed, participants gave short, confirmatory responses such as “mm-hmm” or “No. Not really.” The LLM judgment rated these responses as lower across most dimensions than humans did, indicating a possible response length bias. In 19 of the 40 high-disagreement cases we reviewed, participants use language that is specific to their culture or context without including additional explanation (e.g., Marines referring to "EPWs").

\section{Mixed-Effects Model}
\subsection{Model Equation}
\label{sec:mixed-model-equation}

Because our data has a nested structure where multiple responses come from a single participant and multiple participants come from a single research project, we cannot assume independence between responses. To account for this, we use a linear mixed-effects model given by Equation~\ref{eq:mixed_model}:

\begin{equation}
\label{eq:mixed_model}
Y_{ijk} = \beta_0 + \sum_{p=1}^{10} \beta_p X_{pijk} + u_k + v_{jk} + \epsilon_{ijk}
\end{equation}

$Y_{ijk}$ represents the observed response quality criterion for the $i$-th response provided by the $j$-th participant in the $k$-th research project. $\beta_0$ is the overall fixed intercept of the model. $X_{pijk}$ denotes the value of the $p$-th fixed-effect predictor for a response. The corresponding fixed-effect coefficient, $\beta_p$, captures the relationship between the $p$-th predictor and the response quality. To model the nested variance, $u_k$ represents the random intercept for the $k$-th project, accounting for differences in projects. Similarly, $v_{jk}$ represents the random intercept for the $j$-th participant nested within the $k$-th project, accounting for differences in participants. Finally, $\epsilon_{ijk}$ is the residual error capturing the remaining unexplained variance for each response.

\subsection{Multicollinearity and Variance Inflation}
\label{sec:multicollinearity}

\begin{figure*}
    \centering
    \includegraphics[width=1\linewidth]{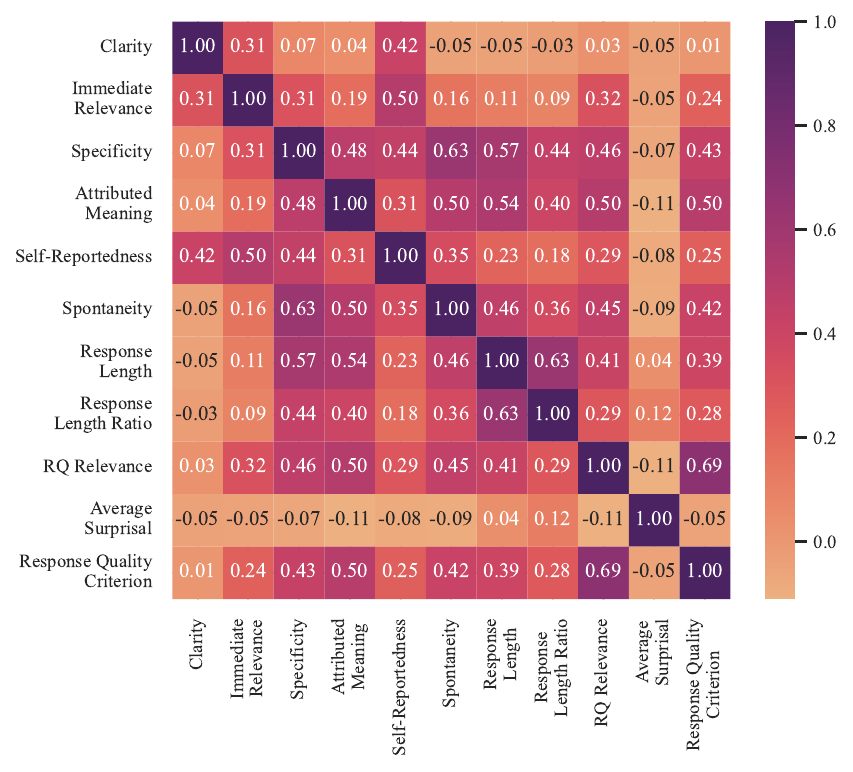}
    \caption{Correlations observed in the Qualitative Interview Corpus between each pair of characteristics in our framework.}
    \label{fig:multicollinearity}
\end{figure*}

We design our framework with distinct characteristics of participant responses to minimize multicollinearity and ensure stable coefficients in our regression. This choice results in low variance inflation factors, which support the stability and interpretability of our mixed-effects model's coefficients (\tref{tab:variance-inflation}). The full correlation among all predictors and our quality criterion is provided in \fref{fig:multicollinearity}.

\subsection{Prompt and Model Selection}
\label{sec:prompt-and-model-selection}

To test whether the results of our mixed-effect regression are robust to model and prompt selection, we relabel the full dataset with three variations of the LLM judges. We use Gemini 3.1 Pro with the original prompts and GPT-5.4 with two AI-generated prompt variations. Using the labels from the LLM judge variations, we rerun the above regression and find that our results are consistent (\tref{tab:regression-robustness}). In all runs, research question relevance is by far the most important predictor. There is some variation in the smaller coefficient predictors, but in 3 of 4 runs clarity is statistically insignificant, and in 3 of 4 runs average surprisal is statistically insignificant. These results demonstrate the robustness of our findings across differences in model and prompt selection.

\subsection{Attenuation Bias}
\label{sec:attenuation-bias}
Because standard mixed-effects regressions do not account for measurement error, and some of our variables rely on LLM judges, it is possible for our coefficients to be biased towards zero. To test whether this attenuation bias affects our results, we run a Simulation Extrapolation (SIMEX) procedure where we inject normally distributed noise into our LLM judge measures according to their reliability as measured by Krippendorff's alpha. We use a 500-sample bootstrap with replacement to estimate the standard errors of the extrapolated estimates. Again, we find that these results are consistent with the original regression (\Cref{tab:simex}). Though some characteristics are no longer statistically significant, research question relevance is still by far the most important predictor, and clarity and average surprisal are still statistically insignificant.

\begin{table*}[t]
\small
\centering
\begin{tabular}{l c c c c c}
\toprule
\textbf{Characteristic} & \textbf{Coef.} & \textbf{Std. Coef.} & \textbf{CI} & \textbf{Std. CI} & \textbf{P-Value} \\ \midrule
\shortstack[l]{Research Question Relevance} & \textbf{0.922}  & \textbf{0.536}  & \textbf{0.900 -- 0.944}    & \textbf{0.523 -- 0.549}    & \textbf{<0.001} \\ \midrule
Attributed Meaning           & \textbf{0.210}  & \textbf{0.137}  & \textbf{0.189 -- 0.230}    & \textbf{0.124 -- 0.150}    & \textbf{<0.001} \\ \midrule
Specificity                  & \textbf{0.097}  & \textbf{0.056}  & \textbf{0.071 -- 0.124}    & \textbf{0.041 -- 0.071}    & \textbf{<0.001} \\ \midrule
Response Length              & \textbf{0.000}  & \textbf{0.048}  & \textbf{0.000 -- 0.000}    & \textbf{0.032 -- 0.064}    & \textbf{<0.001} \\ \midrule
Immediate Relevance          & \textbf{0.081}  & \textbf{0.039}  & \textbf{0.055 -- 0.106}    & \textbf{0.027 -- 0.051}    & \textbf{<0.001} \\ \midrule
Spontaneity                  & \textbf{0.093}  & \textbf{0.037}  & \textbf{0.059 -- 0.128}    & \textbf{0.023 -- 0.051}    & \textbf{<0.001} \\ \midrule
Self-reportedness            & \textbf{0.030}  & \textbf{0.016}  & \textbf{0.004 -- 0.056}    & \textbf{0.002 -- 0.029}    & \textbf{0.037}  \\ \midrule
Average Surprisal            & 0.013           & 0.009           & -0.003 -- 0.030            & -0.002 -- 0.020            & 0.127           \\ \midrule
Response Length Ratio        & 0.000           & 0.002           & -0.002 -- 0.002            & -0.012 -- 0.016            & 0.859           \\ \midrule
Clarity                      & 0.002           & 0.001           & -0.032 -- 0.036            & -0.011 -- 0.012            & 0.912           \\ \bottomrule
\end{tabular}
\caption{Full mixed-effect regression results including  coefficients, standardized coefficients, confidence intervals, and standardized confidence intervals. Bold indicates $p<0.05$.}
\label{tab:full-regression}
\end{table*}

\begin{table*}[t]
\small
\centering
\begin{tabular*}{\textwidth}{@{\extracolsep{\fill}} l c c c c c c c c @{}}
\toprule
 & \multicolumn{2}{c}{\textbf{Original}} & \multicolumn{2}{c}{\textbf{Gemini}} & \multicolumn{2}{c}{\textbf{Prompt Var 1}} & \multicolumn{2}{c}{\textbf{Prompt Var 2}} \\
\cmidrule(lr){2-3} \cmidrule(lr){4-5} \cmidrule(lr){6-7} \cmidrule(lr){8-9}
\textbf{Characteristic} & \textbf{Std. Coef.} & \textbf{P-Value} & \textbf{Std. Coef.} & \textbf{P-Value} & \textbf{Std. Coef.} & \textbf{P-Value} & \textbf{Std. Coef.} & \textbf{P-Value} \\ \midrule
\shortstack[l]{Research Question\\Relevance} & \textbf{0.536} & \textbf{<0.001} & \textbf{0.494} & \textbf{<0.001} & \textbf{0.538} & \textbf{<0.001} & \textbf{0.535} & \textbf{<0.001} \\ \midrule
Attributed Meaning           & \textbf{0.137} & \textbf{<0.001} & \textbf{0.051} & \textbf{<0.001} & \textbf{0.112} & \textbf{<0.001} & \textbf{0.106} & \textbf{<0.001} \\ \midrule
Specificity                  & \textbf{0.056} & \textbf{<0.001} & \textbf{0.058} & \textbf{<0.001} & \textbf{0.099} & \textbf{<0.001} & \textbf{0.063} & \textbf{<0.001} \\ \midrule
Response Length              & \textbf{0.048} & \textbf{<0.001} & \textbf{0.075} & \textbf{<0.001} & \textbf{0.032} & \textbf{0.001}  & \textbf{0.117} & \textbf{<0.001} \\ \midrule
Immediate Relevance          & \textbf{0.039} & \textbf{<0.001} & \textbf{0.032} & \textbf{<0.001} & \textbf{0.055} & \textbf{<0.001} & 0.022          & 0.073          \\ \midrule
Spontaneity                  & \textbf{0.037} & \textbf{<0.001} & 0.011          & 0.186          & \textbf{0.040} & \textbf{<0.001} & \textbf{0.048} & \textbf{<0.001} \\ \midrule
Self-reportedness            & \textbf{0.016} & \textbf{0.037}  & \textbf{0.057} & \textbf{<0.001} & 0.000        & 0.982          & \textbf{-0.039}& \textbf{0.006}  \\ \midrule
Average Surprisal            & 0.009          & 0.127          & 0.006          & 0.375          & 0.002          & 0.761          & \textbf{-0.044}& \textbf{<0.001} \\ \midrule
Response Length Ratio        & 0.002          & 0.859          & -0.013        & 0.162          & -0.004        & 0.652          & 0.005          & 0.717          \\ \midrule
Clarity                      & 0.001          & 0.912          & 0.008          & 0.238          & \textbf{0.020} & \textbf{0.007}  & 0.013          & 0.250          \\ \bottomrule
\end{tabular*}
\caption{Comparison of the regression analysis across different model and prompt selections. Original indicates GPT-5.4 with the original prompts. Gemini indicates Gemini 3.1 Pro with the original prompts. Prompt Var 1 indicates GPT-5.4 with an AI-generated prompt variation. Prompt Var 2 indicates GPT-5.4 with a different AI-generated prompt variation. Bold indicates $p<0.05$.}
\label{tab:regression-robustness}
\end{table*}

\section{Interview Techniques}
\subsection{Technique Taxonomy}
We use \citeposs{kvale_interviews_2009} taxonomy of interview techniques to conduct our analysis (\tref{tab:techniques}).

\subsection{Dunn's Post-hoc Test}
\label{sec:dunns-test}

\begin{figure*}
    \centering
    \includegraphics[width=1\linewidth]{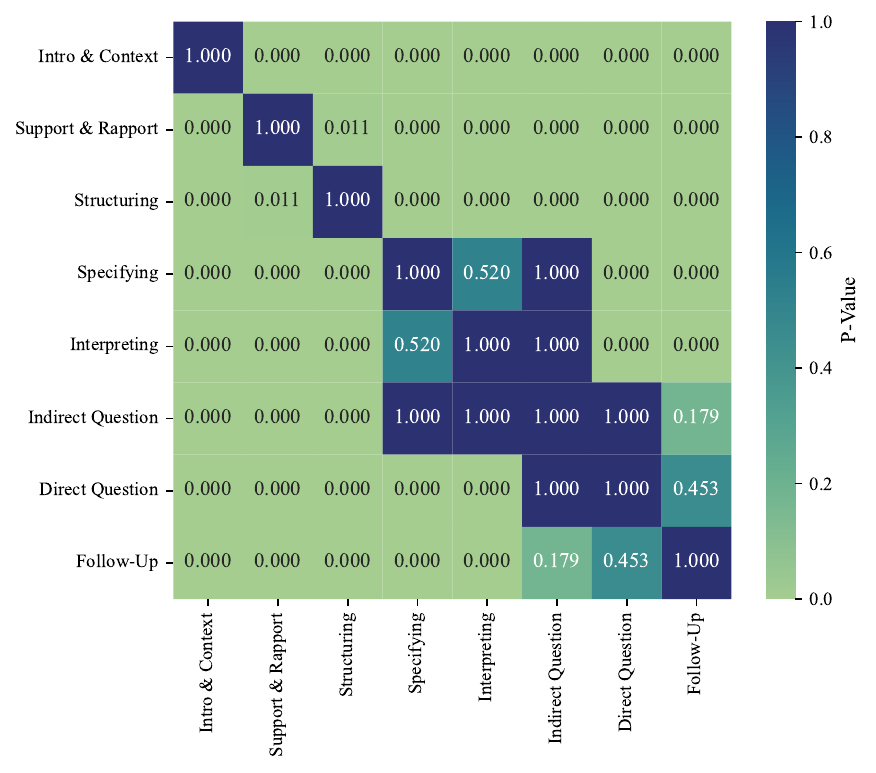}
    \caption{P-values from Dunn's post-hoc test for difference in median response quality between pairs of interview techniques.}
    \label{fig:dunns-test}
\end{figure*}

In \sref{sec:techniques-analysis}, we identify techniques used in interview excerpts and use Dunn's post-hoc test with Bonferroni correction to identify statistically significant differences in medians between pairs of techniques. \fref{fig:dunns-test} shows the full set of p-values for Dunn's post-hoc test.

\end{document}